\documentclass{article}
\usepackage{spconf,amsmath,graphicx}
\usepackage{hyperref}
\usepackage{multirow}
\usepackage{graphicx}

\usepackage{enumitem}
\setlist{nosep, leftmargin=14pt}

\usepackage{mwe} 

\usepackage{amsfonts,amssymb}

\usepackage{color}

\def\ie{\emph{i.e.}}

\title{MeLo: Low-rank Adaptation is Better than Fine-tuning for Medical Image Diagnosis}
\name{\parbox{\linewidth}{\centering Yitao Zhu$^{1}$\qquad Zhenrong Shen$^{2}$\qquad Zihao Zhao$^{1}$ \qquad Sheng Wang$^{1,2,3}$ \\
Xin Wang$^{2}$ \qquad Xiangyu Zhao$^{2}$ \qquad Dinggang Shen$^{1, 3, 4}$ \qquad Qian Wang$^{1,4}$}}
\address{$^1$School of Biomedical Engineering \& State Key Laboratory of \\Advanced Medical Materials and Devices, 
ShanghaiTech University, Shanghai, 201210, China \\
$^2$School of Biomedical Engineering, Shanghai Jiao Tong University, Shanghai, 200030, China \\
$^3$Shanghai United Imaging Intelligence Co., Ltd., Shanghai, 200230, China \\
$^4$Shanghai Clinical Research and Trial Center, Shanghai, 201210, China}
\begin{document}
%
\maketitle
\begin{abstract}
The common practice in developing computer-aided diagnosis (CAD) models based on transformer architectures usually involves fine-tuning from ImageNet pre-trained weights. However, with recent advances in large-scale pre-training and the practice of scaling laws, Vision Transformers (ViT) have become much larger and less accessible to medical imaging communities. 
Additionally, in real-world scenarios, the deployments of multiple CAD models can be troublesome due to problems such as limited storage space and time-consuming model switching.
To address these challenges, we propose a new method MeLo (\textbf{Me}dical image \textbf{Lo}w-rank adaptation), which enables the development of a single CAD model for multiple clinical tasks in a lightweight manner. It adopts low-rank adaptation instead of resource-demanding fine-tuning. 
By fixing the weight of ViT models and only adding small low-rank plug-ins, we achieve competitive results on various diagnosis tasks across different imaging modalities using only a few trainable parameters. 
Specifically, our proposed method achieves comparable performance to fully fine-tuned ViT models on four distinct medical imaging datasets using about 0.17\% trainable parameters. 
Moreover, MeLo adds only about 0.5MB of storage space and allows for extremely fast model switching in deployment and inference.
Our source code and pre-trained weights are available \href{https://absterzhu.github.io/melo.github.io/}{\textbf{here}}.
\end{abstract}

\begin{figure}[t]
    \centering
    \includegraphics[width=0.48\textwidth]{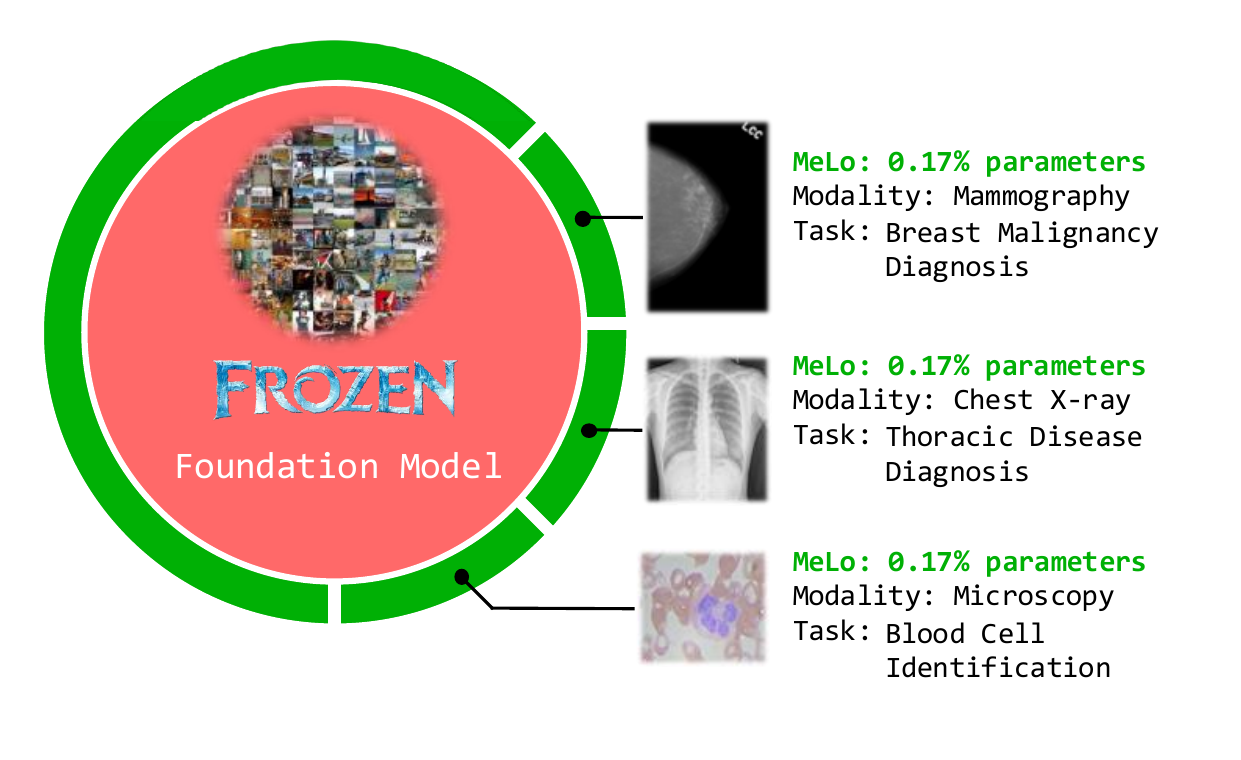}
    \caption{The motivation of MeLo. The large-scale vision foundation model is just like a watermelon, and our proposed MeLo can conveniently adjust it to different clinical tasks by few additional parameters. 
    }
    \label{overview}
\end{figure}

\begin{figure*}[t]
    \centering
    \includegraphics[width=0.95\textwidth]{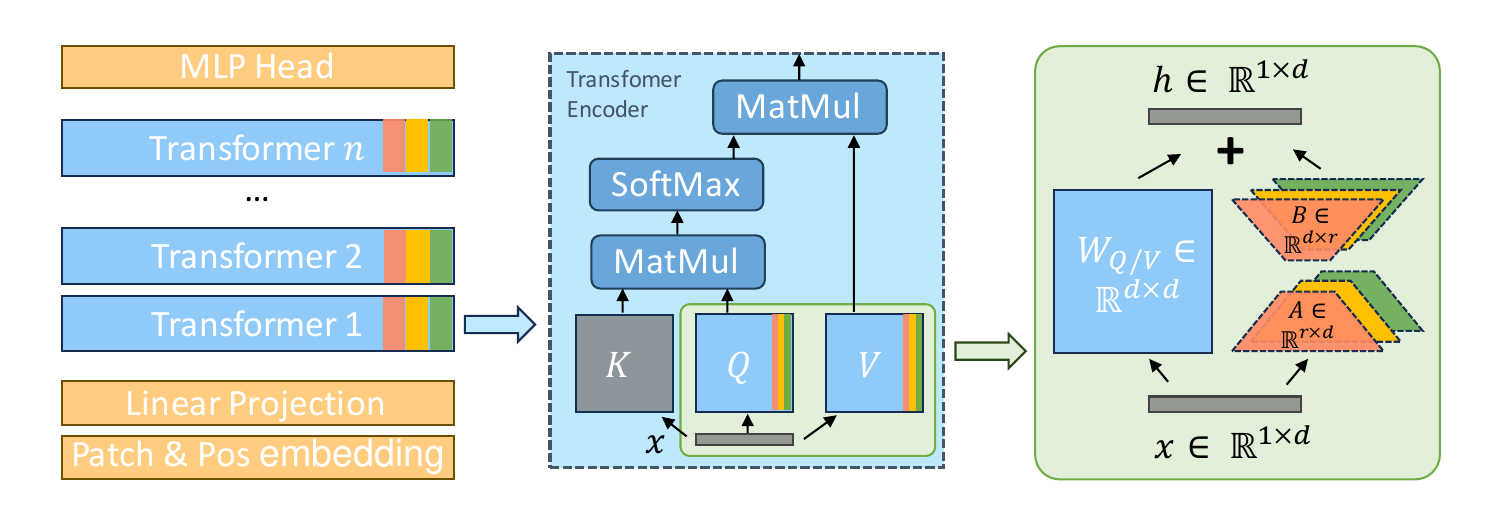}
    \caption{The illustration of our proposed MeLo. For a specific medical image diagnosis task, we inject low-rank decomposition matrices (denoted as $A$ and $B$) into the pre-trained query and value projection matrices (denoted as $W_Q$ and $W_V$) of each self-attention layer. Different module colors respond to different clinical tasks.}
    \label{method}
\end{figure*}

\section{Introduction}
\label{sec:intro}
In the last decade, deep learning in computer vision has undergone a revolution and significantly impacted the field of medical image analysis. 
Especially Vision Transformer (ViT)~\cite{dosovitskiy2020image} has demonstrated remarkable capabilities in learning complex representations in a data-driven manner. 
However, training ViTs from scratch typically requires large annotated datasets, which are challenging to collect in healthcare due to privacy concerns and expensive annotation~\cite{razzak2018deep}. 
Thus, transfer learning by leveraging pre-trained weight on ImageNet~\cite{deng2009imagenet} has gained popularity by serving as a warm starting point. 
More recently, the community has witnessed an increase in model scale~\cite{cherti2023reproducible,chen2022scaling}, especially ViT\cite{dehghani2023scaling} which exhibits superior generalizability and robustness compared to smaller ImageNet pre-trained ViT when transferred to other domains. 
Thus, the success of ViT and its variants, which are typically called visual foundation models, has inspired subsequent research on medical foundational models aiming to build more powerful CAD systems.

Although these vision foundation models have shown unprecedented capabilities, their deployment and maintenance present several challenges in real clinical scenarios.
Constantly updating images from new devices and new epidemics means frequent fine-tuning, which is time-consuming and resource-intensive since the models are extremely large. 
Moreover, in clinical practice, a paramount consideration resides in the optimization of storage space utilization, the reduction of GPU resource demands, and the expeditious execution of diverse medical image processing tasks. Nonetheless, the process of fine-tuning multiple foundational models \textit{not only} imposes substantial storage space overhead \textit{but also} exacerbates latency due to frequent model switching on the GPU.


To address the challenges above, we propose MeLo (\textbf{Me}dical image \textbf{Lo}w-rank adaptation), a novel approach that leverages low-rank adaptation instead of full fine-tuning to efficiently transfer a pre-trained vision foundation model to a powerful CAD. 
MeLo freezes the original weights of visual foundation models while adding small low-rank plug-ins that can achieve remarkable results using only a small fraction of trainable weights.
Trained with less than 0.175\% of original parameters, the ViT-based models can achieve performance comparable to the fully fine-tuned counterparts. 
This substantial reduction in trainable parameters translates to a significant reduction in computational resources and training time, making it more feasible for researchers and practitioners to deploy and maintain high-performing models.
Since MeLo is a very small plugin, a ViT-based foundation model and the MeLo for multiple tasks can be loaded all at once, and the corresponding MeLo module can be activated when dealing with a specific diagnosis task, so as to achieve fast task response.

To demonstrate the effectiveness and versatility of our proposed MeLo, we have conducted extensive experiments on various diagnosis tasks across different imaging modalities. 
In each of these tasks, MeLo consistently matches or even outperforms the performance of fully fine-tuned models, while utilizing significantly fewer trainable parameters. 
We also test the deployment and inference of MeLo when confronting multitasking, and demonstrate that MeLo \textit{not only} significantly alleviates the demand for GPU memory \textit{but also} facilitates expedited switching between different tasks.

In summary, the proposed MeLo offers a powerful and efficient alternative to the fully fine-tuned methods for medical image analysis. 
By utilizing a small fraction of the trainable parameters and preserving pre-trained weights, MeLo enables researchers and practitioners to develop high-performing and robust models that can be easily deployed and maintained in a wide range of medical imaging applications.
We are committed to making MeLo widely available to the research and clinical communities. 
To this end, we have made the MeLo system publicly accessible, along with pre-trained MeLo module weights. 
With these lightweight weights, users can obtain a diagnosis model that boasts similar or superior performance to their fine-tuned counterparts while benefiting from the reduced complexity and resource requirements.

\section{method}
\label{sec:method}
In this section, we will first present the methodology of the proposed MeLo in detail, and then introduce the datasets and the implementation details in our experiments.

\begin{table*}[htbp]
\centering
\caption{Performance comparison of fine-tuning and MeLo on three medical image diagnosis tasks using four different datasets.}
\resizebox{\textwidth}{!}{
\begin{tabular}{ccccccccccc}
\hline
Dataset                    & Task            & Classes  & Method               & Parameters & Trainable & ACC            & SEN            & PRE            & F1S            & AUC            \\ \hline
\multirow{2}{*}{\textbf{Shenzhen Hospital Chest X-ray}}  & \multirow{2}{*}{Tuberculosis Diagnosis} & \multirow{2}{*}{2} & Fine-tuning & 81.825M    & 81.825M   & 0.812          & 0.813          & 0.813          & 0.812          & 0.894          \\
                           &                    &                           & MeLo        & 81.966M    & 0.142M    & \textbf{0.835} & \textbf{0.833} & \textbf{0.836} & \textbf{0.834} & \textbf{0.898} \\ \hline
\multirow{2}{*}{\textbf{BloodCell}} & \multirow{2}{*}{Blood Cell Identification} & \multirow{2}{*}{4} & Fine-tuning & 81.827M    & 81.827M   & 0.859          & \textbf{0.877}          & 0.888          & 0.868          & \textbf{0.983} \\
                           &                    &                           & MeLo        & 81.968M    & 0.144M    & \textbf{0.930} & 0.875 & \textbf{0.958} & \textbf{0.910} & 0.947          \\ \hline
\multirow{2}{*}{\textbf{INbreast}}  & \multirow{2}{*}{Breast Malignancy Diagnosis} & \multirow{2}{*}{2} & Fine-tuning & 81.825M    & 81.825M   & \textbf{0.748} & 0.594          & \textbf{0.674} & 0.554          & 0.684          \\
                           &                    &                           & MeLo        & 81.966M    & 0.142M    & 0.745          & \textbf{0.604} & 0.615          & \textbf{0.572} & \textbf{0.687} \\ \hline
\multirow{2}{*}{\textbf{NIH Chest X-ray14}}  & \multirow{2}{*}{Thoracic Disease Diagnosis} & \multirow{2}{*}{14} & Fine-tuning & 81.825M    & 81.825M   & \textbf{0.369} & 0.094          & 0.316 & 0.132          & 0.788          \\
                           &                    &                           & MeLo        & 81.980M    & 0.157M    & 0.357          & \textbf{0.106} & \textbf{0.319}          & \textbf{0.142} & \textbf{0.794} \\ \hline
\end{tabular}}
\label{tab:LoRA_3datasets}%
\end{table*}

\begin{table*}[htbp]
  \centering
  \caption{Performance on 100 images from four datasets (25 per dataset) in dataset-specific order and random order. IT is the total time for model initialization, ST is the total time for model switching, A-ST is the average time per switch, TT is the total inference time for all images, and Parameter is the total number of model parameters. Fine-tune$^{*}$ loads four fine-tuned ViT models for each dataset into GPU memory all at once.}
    \begin{tabular}{c|cccc|cccc|c}
\hline
\multirow{2}{*}{Method} & \multicolumn{4}{c|}{In Order}                                         & \multicolumn{4}{c|}{Random}                                            & \multirow{2}{*}{Parameter} \\ \cline{2-9}
                        & IT              & ST              & A-ST            & TT              & IT               & ST              & A-ST            & TT              &                       \\ \hline
Fine-tuing              & \textbf{25.9s} & 11.5s         & 2.9s          & 19.6s         & \textbf{25.4s} & 214.0s        & 2.9s          & 218.5s        & 1758.6M              \\
Fine-tuing$^*$          & 118.1s        & $<$0.1s          & $<$0.1s            & \textbf{7.9s} & 121.3s         & $<$0.1s               & $<$0.1s               & \textbf{8.2s} & 7034.4M              \\
MeLo                    & 34.0s         & $<$0.1s & $<$0.1s & 8.9s          & 34.4s          & $<$0.1s & $<$0.1s & 10.6s         & 1759.9M      \\ \hline
\end{tabular}%
\label{tab:computation}%
\end{table*}%

\subsection{Medical Image Low-rank Adaptation (MeLo)}
To efficiently and effectively transfer a vision foundation model to a specific CAD model, we turn to Low-Rank Adaptation (LoRA)~\cite{hu2021lora}, a popular parameter-efficient fine-tuning technique, to build our proposed MeLo method in this study.
LoRA was first proposed to fine-tune large language models according to the hypothesis that the weight change of a pre-trained large model is highly sparse and has a low intrinsic rank during fine-tuning. 
The main idea of LoRA is to freeze the pre-trained model weights and inject trainable rank decomposition matrices into each layer of the Transformer architecture, thus greatly reducing the number of trainable parameters when fine-tuning large models. 
A lot of studies turn to LoRA instead of full fine-tuning for building their own large models without having access to intensive computational resources.
As for MeLo, we employ LoRA to efficiently adapt ViT based models to different diagnosis applications as illustrated in Figure~\ref{overview}.

For a specific clinical scenario, we add LoRA weights into each self-attention layer of a pre-trained ViT as depicted in Figure~\ref{method}.
For the pre-trained query and value projection matrices (denoted as $W_Q$ and $W_V$) in a self-attention layer,
the added LoRA weights constrain their updates by representing them with a low-rank decomposition during fine-tuning, which can be expressed as:
\begin{equation}
    h = W_{0}x + \triangle Wx = W_{0}x + BAx
\end{equation}
\noindent where $x\in\mathbb{R}^{1\times d}$ and $h\in\mathbb{R}^{1\times d}$ stand for the input and output features, respectively.
Two low-rank matrices, $B \in \mathbb{R}^{d\times r}$ and $A\in \mathbb{R}^{r\times d}$, compose the weight change $\triangle W$ of the pre-trained weight $W_{0}$.
The ranks $r$ of these low-rank matrices are much smaller than the model dimension $d$, and we empirically set $r=4$ in our experiments.
By switching different MeLo modules, one pre-trained ViT can effectively handle different medical image diagnosis tasks, thus greatly reducing the computation budget for building a versatile CAD system especially when the training data is limited.

\subsection{Datasets and Implementation Details}
To comprehensively evaluate the utility of MeLo, we conduct experiments using four datasets for three different medical image diagnosis tasks including thoracic disease diagnosis in chest X-ray (CXR) images, breast malignancy diagnosis in mammography images, and blood cell identification in microscopic slides, which are described below.

\begin{itemize}


\item \textbf{Shenzhen Hospital Chest X-ray dataset} was collected by Shenzhen No.3 Hospital in Shenzhen, Guangdong Providence, China. It consists of 326 normal CXR images and 336 abnormal CXR images showing various manifestations of tuberculosis. Our task is to diagnose whether a CXR image displays tuberculosis. We randomly split the dataset with 80\% images for training and 20\% for test. 
The dataset would be available \href{https://lhncbc.nlm.nih.gov/LHC-downloads/downloads.html#tuberculosis-image-data-sets}{\textbf{here}}.

\item \textbf{NIH Chest X-ray 14 dataset}~\cite{wang2017chestx}
comprises 112,120 frontal-view CXR images annotated with 14 common thoracic diseases, and our task is to diagnose the diseases contained in each CXR image. We randomly allocated 70\% of the images for training, 10\% for validation, and the remaining 20\% for testing. 

\item \textbf{INBreast dataset}~\cite{InsMoreira2012INbreastTA} includes 410 digital mammography images from 115 patients, which consists of 339 non-malignant and 71 malignant ones. 
The diagnosis task follows BI-RADS assessment of masses~\cite{LauraLiberman2002BreastIR} to classify these mammography images into non-malignant and malignant ones.
We randomly split the dataset with 80\% images for training and 20\% for testing. 


\item \textbf{BloodCell dataset} contains 12,500 augmented images of four blood cell subtypes including Eosinophil, Lymphocyte, Monocyte, and Neutrophil. Our task is to identify their blood cell types. We used the dataset's own partitioning and the dataset would be available \href{https://www.kaggle.com/datasets/paultimothymooney/blood-cells/data}{\textbf{here}}.

\end{itemize}
 
All models in experiments were trained by ourselves. During training, we use the learning rate of $3\times10^{-4}$ and Adam optimizer to train MeLo based on ViTs for 200 epochs until convergence, and save the weights with the best validation performance as the final test model.
In Experiment \ref{exp1}, we use the ViT-base model pre-trained on ImageNet. 
In Experiment \ref{exp2}, we turn to ViTs pre-trained with CLIP~\cite{radford2021learning} with varying model sizes, including ViT-base, ViT-huge, and ViT-giga models. 
In Experiment \ref{exp3}, we use the ViT-giga model pre-trained with CLIP for evaluation. 
The pre-trained weights of all ViT models used in our experiments are provided in~\cite{rw2019timm}.
All the experiments are conducted using a single Nvidia A100 80G GPU.

\section{Experiments}
\label{sec:experiments}

\subsection{Performance on Different Diagnosis Tasks}
\label{exp1}
We test image classification performance on MeLo using four distinct datasets comprising various modalities of medical images: \textbf{chest X-rays}, \textbf{blood smears}, and \textbf{mammograms}. The experimental results are presented in Table~\ref{tab:LoRA_3datasets}.
It is evident that MeLo demonstrates either similar or improved performance across all datasets in various evaluation metrics.
In contrast to the fully fine-tuned which trains 81 million parameters, MeLo has significantly fewer trainable parameters, \ie, 0.14 million. 
These experiments affirm that MeLo exhibits superior efficiency compared to full fine-tuning approaches.


\subsection{Performance on Different ViT Models}
\label{exp2}
\begin{figure}
    \centering
    \includegraphics[width=0.48\textwidth]{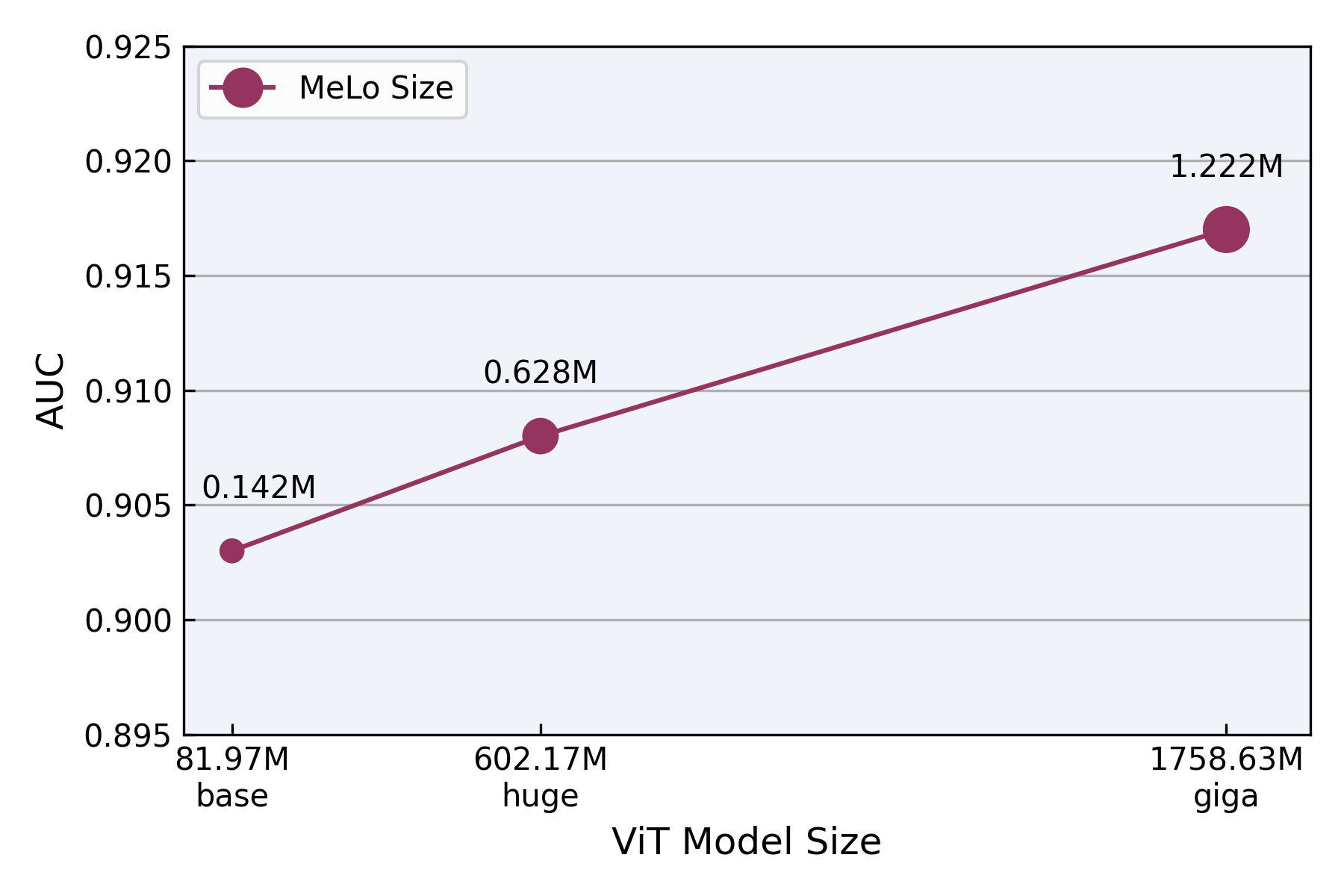}
    \caption{The AUC gradually increases as the ViT model size expands while the trainable parameters of corresponding MeLo modules remain consistently low.}
    \label{model_size}
\end{figure}

We test the effectiveness of MeLo on ViT models with varying sizes using the Shenzhen Hospital X-ray dataset.
The results in Figure~\ref{model_size} show that the AUC demonstrates an increment as the ViT model size increases. 
Moreover, while the model capacities of ViTs increase substantially with scale, their MeLo module maintains a consistently low number of trainable parameters across model sizes.
For example, there are only 1.22 million trainable parameters in MeLo for the giga ViT model which comprises 1759 million parameters. 
These findings indicate that in the future, as the pre-trained ViTs continue to expand their model sizes, performance improvement can still be achieved at a minimal cost in terms of trainable parameters.

\subsection{Performance on Deployment and Inference}
\label{exp3}
To comprehensively evaluate the deployment and inference of our proposed MeLo, we conduct a simulation experiment to verify its effectiveness.
We first collect 100 medical images of varying imaging modalities by randomly selecting 25 images from each dataset.
Then two data processing situations are simulated, i..e, one situation in which the images from a specific dataset are sequentially processed one by one, and the other situation in which all the images from different datasets are shuffled and processed in a totally random order.
We load a ViT-giga model and four MeLo modules for different diagnosis tasks together, and compare this deployment with four ViT-giga models that are fully fine-tuned for each corresponding dataset.
There are two deployment strategies for the fully fine-tuned models, , i..e, one is to temporarily load the corresponding model when encountering a specific task, and the other is to preload all models at once.
The experimental results in Table~\ref{tab:computation} illustrates that MeLo provides significant latency and memory optimization benefits during deployment and inference.
Specifically, the model equipped with MeLo leads to lower inference time and smaller GPU memory usage when dealing with multiple diagnosis tasks in different orders.
As for the fully fine-tuned models deployed in the second strategy (2nd row in Table~\ref{tab:computation}), it is worth noting that the clinical practice typically involves dozens, if not hundreds, of different diagnostic tasks where a large number of models are supposed to be deployed.
Such a deployment strategy would definitely lead to GPU memory explosion, making it infeasible in a real-world scenario.

\section{CONCLUSION AND DISCUSSION}
In this work, we propose MeLo, a highly efficient and easily accessible approach that leverages low-rank adaptation to transfer a pre-trained vision foundation model to a versatile CAD system.
MeLo has been evaluated across various medical image diagnosis tasks using different imaging modalities and ViT models with different sizes.
The experiments consistently prove its similar or superior performance compared with full fine-tuning while utilizing significantly fewer trainable parameters and showing the superiority of latency and memory usage in practical deployments. 
The outstanding performance highlights the potential to establish a community focused on creating high-performing, robust, and equitable models that can be readily deployed and maintained across a broad spectrum of medical image diagnosis applications in a short timeframe.

\section{COMPLIANCE WITH ETHICAL STANDARDS}
This research study was conducted retrospectively using human subject data made available in open access. Ethical approval was not required as confirmed by the license attached with the open access data.


\bibliographystyle{IEEEbib}
\bibliography{strings,refs}

\end{document}